\documentclass[11pt,a4paper]{article}
\usepackage[hyperref]{emnlp2020}
\usepackage{times}
\usepackage{latexsym}

\usepackage{graphicx}
\usepackage{amsmath}
\usepackage{amsthm}
\usepackage{amsfonts}
\usepackage{booktabs}
\usepackage{algorithm}
\usepackage{algorithmic}
\usepackage{subcaption}
\usepackage{multirow}
\usepackage{pifont}
\def \MethodName {{\textsc IBT}}

\usepackage{bm}

\usepackage{tikz}
\usepackage{pgfplots}
\pgfplotsset{width=5cm,compat=1.3, legend style={at={(0.40,0.40)},anchor=north west}}

\usepackage{microtype}

\aclfinalcopy 
\def\aclpaperid{737} 

\title{A Simple Baseline to
Semi-Supervised 
Domain Adaptation for \\
Machine Translation}

\author{Di Jin,\textsuperscript{\rm 1} Zhijing Jin,\textsuperscript{\rm 2} Joey Tianyi Zhou,\textsuperscript{\rm 3} Peter Szolovits\textsuperscript{\rm 1}\\  
\textsuperscript{\rm 1}Computer Science \& Artificial Intelligence Laboratory, MIT \\
\textsuperscript{\rm 2}Amazon Web Services
\\
\textsuperscript{\rm 3}A*STAR, Singapore
\\
\texttt{\{jindi15,psz\}@mit.edu}\\
\texttt{zhijing.jin@connect.hku.hk}\\
\texttt{zhouty@ihpc.a-star.edu.sgu}
} 

\date{}

\begin{document}
\maketitle
\begin{abstract}
State-of-the-art neural machine translation (NMT) systems are data-hungry and perform poorly on new domains with no supervised data. As data collection is expensive and infeasible in many cases, domain adaptation methods are needed. In this work, we propose a simple but effect approach to the semi-supervised domain adaptation scenario of NMT, where the aim is to improve the performance of a translation model on the target domain consisting of only non-parallel data with the help of supervised source domain data. This approach iteratively trains a Transformer-based NMT model via three training objectives: language modeling, back-translation, and supervised translation.
We evaluate this method on two adaptation settings: adaptation between specific domains and adaptation from a general domain to specific domains, and on two language pairs: German to English and Romanian to English. With substantial performance improvement achieved---up to $+19.31$ BLEU over the strongest baseline, and $+47.69$ BLEU improvement over the unadapted model---we present this method as a simple but tough-to-beat baseline in the field of semi-supervised domain adaptation for NMT.\footnote{Our source code is available at \url{https://github.com/jind11/DAMT}}
\end{abstract}

\section{Introduction}

\begin{figure}[!t]
    \centering
    \includegraphics[width= \columnwidth]{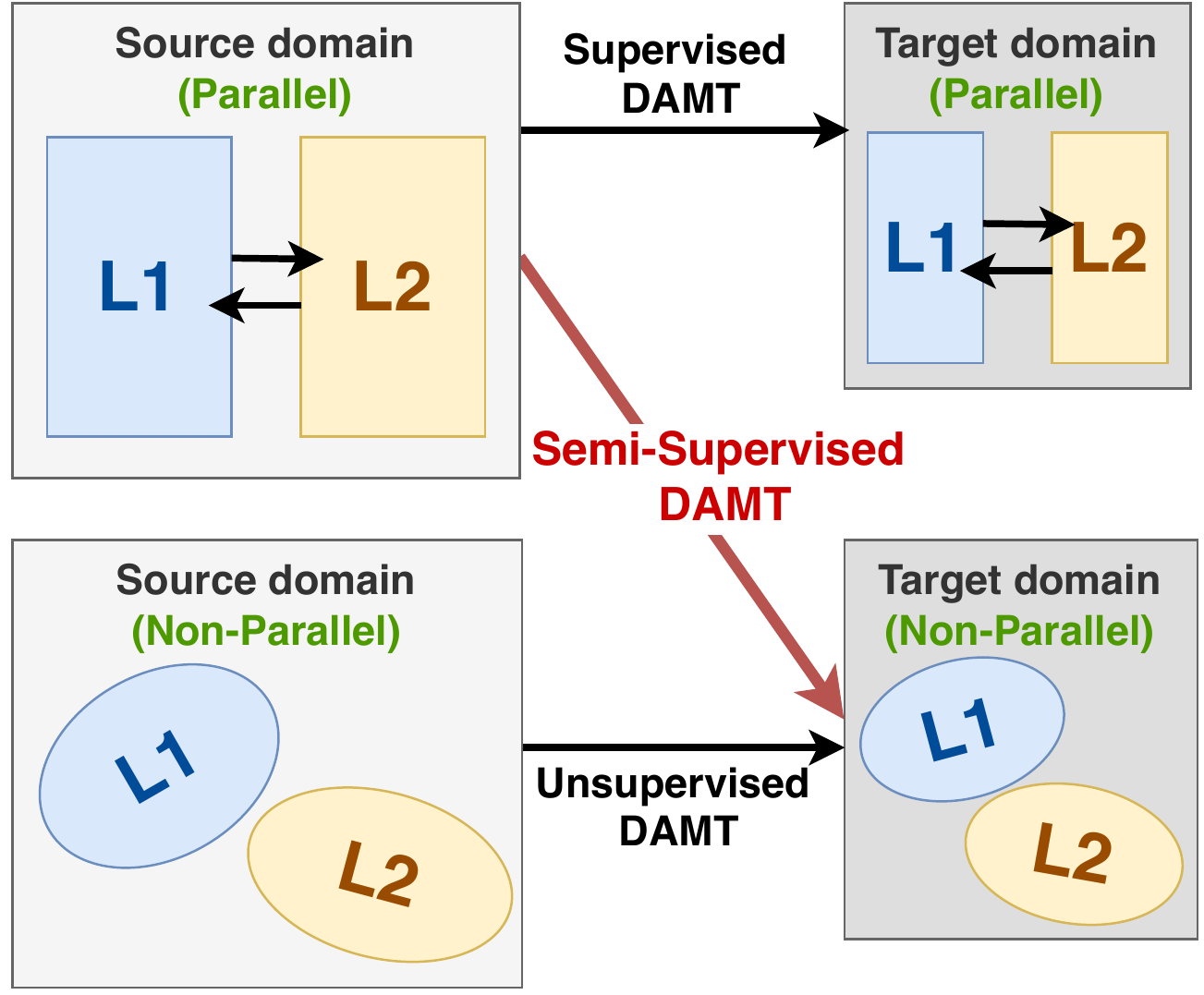}
    \caption{Three types of DAMT: supervised, unsupervised, and semi-supervised (the focus of our paper). L1 is the source language and L2 is the target language of MT.}
    \label{fig:damt_types}
\end{figure}

Machine Translation (MT) is an attractive and successful research field. For many general domains, millions of parallel data are annotated. For example, the WMT14 dataset alone has 4M supervised sentence pairs. However, for more specific domains such as law, medicine, and technology \cite{nakov-tiedemann-2012-combining}, there is very few or hardly any supervised data. In practice, collecting supervised data in specialized fields is expensive and in some cases even impractical.


To obtain a good translation model on these specialized domains (our target domains), semi-supervised domain adaptation for machine translation (DAMT) has become an active research field. It aims to 
generalize the MT models trained on the source domain with large-scale supervised data to the target domain that has no supervised data at all, as illustrated in Figure \ref{fig:damt_types}.
Existing approaches for semi-supervised DAMT can be categorized into two lines: \textit{model-centric} and \textit{data-centric} methods. Model-centric methods focus on multi-task learning of the translation task on the source domain parallel data and the language modeling task on the target domain target-side (target-language) data~\cite{Domhan2017UsingTM,Dou2019UnsupervisedDA}. Whereas, data-centric methods rely on a pseudo-parallel corpus constructed either by simply copying the target-side sentences to the source side in the target domain, termed as \textit{Copy}
~\cite{Currey2017CopiedMD}, or by
pairing the target-side sentences with their translated counterparts by a well-trained MT model, termed as \textit{back-translation}~\cite{sennrich2015improving}. 
Specifically, back-translation first generates translations from the target language to the source language, and then trains the translation model to map the generated sentences back to the target side.
Despite its simplicity, the \textit{back-translation} strategy has been demonstrated to be most effective in many cases. 

\begin{figure}[!htpb]
    \centering
    \includegraphics[width= \columnwidth]{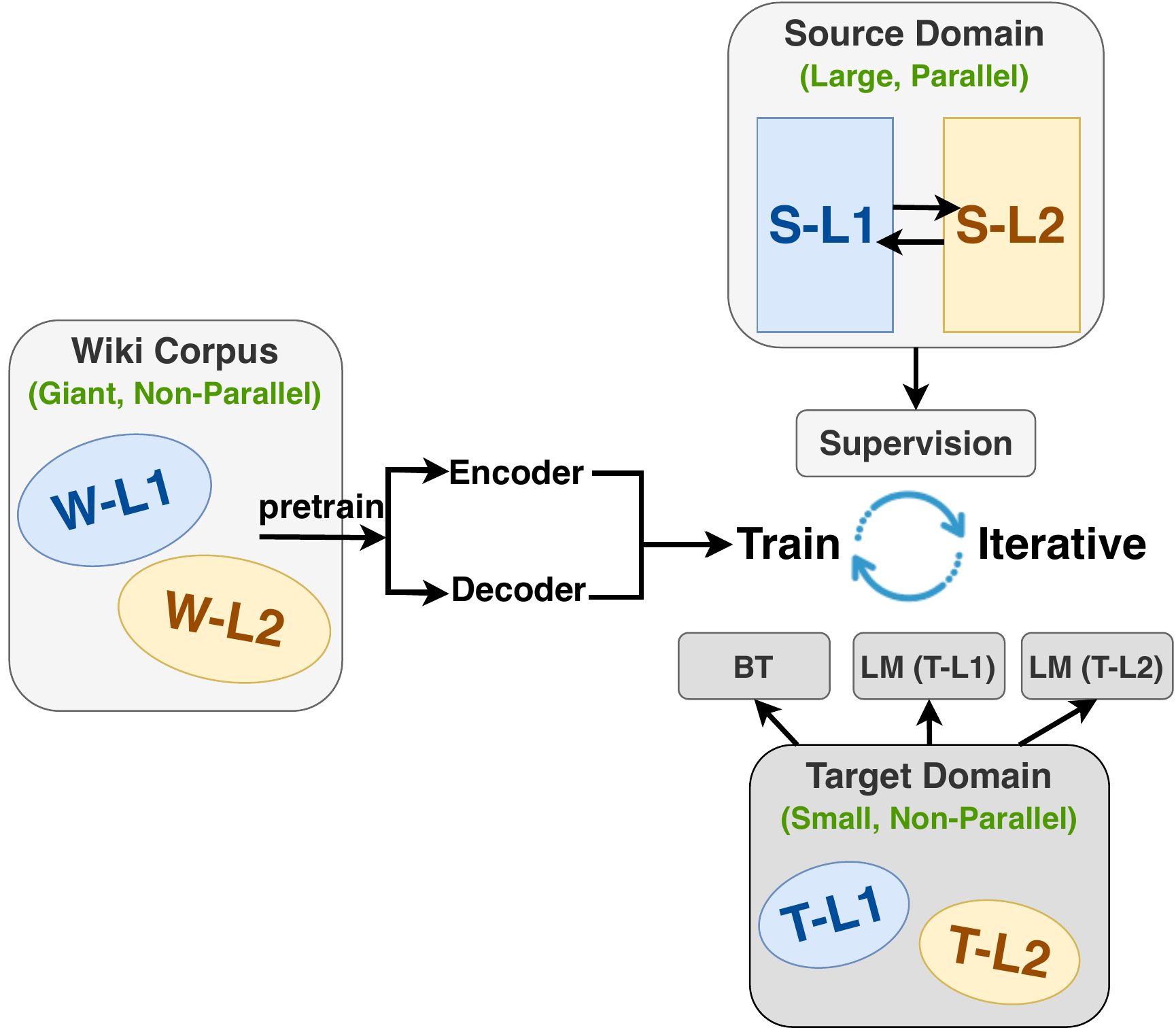}
    \caption{The schema of our method. We first initialize the encoder and decoder via language modeling pre-training over the wiki monolingual corpora for both the source language (W-L1) and target language (W-L2). Then we train the model via iteratively optimizing the back-translation loss (BT in both T-L1$\rightarrow$T-L2 and T-L2$\rightarrow$T-L1 directions) and two language modeling losses (LM (T-L1) and LM (T-L2)) on the non-parallel target domain data, as the base method. We further enhance this base by adding one more optimization step on the supervised translation loss using the source domain data (e.g., S-L1$\rightarrow$S-L2).}
    \label{fig:method-schema}
\end{figure}

Inspired by the success of back-translation, we propose a simple but much stronger approach as illustrated by Figure \ref{fig:method-schema}. We first initialize both encoder and decoder of the sequence-to-sequence model with pre-trained parameters as a good starting point. The pre-training process is implemented via language modeling over large-scale monolingual corpora from Wikipedia. Afterwards, we implement iterative \textit{back-translation} (in both L1$\rightarrow$L2 and L2$\rightarrow$L1 directions) and \textit{language modeling} training over the target domain non-parallel data, which serves as our base method. We further augment this base method by incorporating the supervised translation training step over the source domain data into each iteration, which leads to more significant performance gains. 


Despite the simple nature of this method, we call attention to our approach because it demonstrates \textit{significant} improvements over all previous state-of-the-art DAMT models on all experiments. We conduct experiments
with two different domain adaptation settings and on two language pairs. 
First, for domain adaptation between two specific domains, our base method achieves up to $+9.48$ BLEU score improvement over the strongest out of four baseline models and $+27.77$ BLEU over the unadapted baseline.
Second, for domain adaptation from a general domain with large-scale supervised data (WMT) to specific domains, our model combined with data augmentation by supervised source domain data achieves up to $+19.31$ BLEU improvement over the best previous method and $+47.69$ BLEU improvement over the unadapted model.

\section{Related Work}

There are three scenarios for domain adaptation for machine translation, as illustrated in Figure~\ref{fig:damt_types}: 
\begin{enumerate}
    \item \textbf{Supervised:} Both source and target domains have supervised parallel data, although the data amount of source domain is much larger than the target.
    \item \textbf{Unsupervised:} Neither of the source and target domains has parallel data.
    \item \textbf{Semi-supervised:} Only the source domain has parallel data while the target does not.
\end{enumerate} 

\paragraph{Supervised DAMT} Most previous works for DAMT focus on the supervised setting~\cite{chu-etal-2017-empirical,fadaee-monz-2018-back,guzman-etal-2019-flores}. For example, sequential fine-tuning~\cite{luong2015stanford,freitag2016fast} first trains an neural machine translation (NMT) model on source domain data and subsequently fine-tune it on the target domain data. \citet{britz-etal-2017-effective} proposes to jointly train the translation task and the domain discrimination task to mitigate the domain shift. \citet{Kobus2016DomainCF} uses the domain tokens and domain embeddings to force the NMT model to take into account the domain information. \citet{joty-etal-2015-avoid} assigns higher weights to those source domain data that more resemble the target domain ones so as to remove unwanted noise.

\paragraph{Unsupervised DAMT} DAMT in the unsupervised setting has started only recently, where \citet{sun2019empirical} defines several scenarios for it and proposes modified domain adaptation methods to improve the performance of adaptation in these scenarios.

\paragraph{Semi-Supervised DAMT} Our proposed baseline falls under the semi-supervised scenario, where related works can be divided into two threads: data-centric and model-centric. Data-centric methods mainly propose to select or generate domain-related pseudo-parallel data for model training. For data selection, \citet{duh-etal-2013-adaptation} uses language models to rank the source domain data 
and select the top-ranked parallel sentences as synthetic data. More representative methods are back translation-based~\cite{sennrich2015improving} and copy-based~\cite{Currey2017CopiedMD}, which are simple yet have been widely demonstrated to be effective. 
On the other hand, model-based methods propose to change model architectures to leverage the monolingual corpus by introducing a new learning objective, such as auto-encoding~\cite{cheng-etal-2016-semi} and language modeling~\cite{ramachandran-etal-2017-unsupervised,Dou2019UnsupervisedDA} on the target-side sentences. 

In contrast, we would like to re-visit the classic \textit{back-translation} method and propose extending it to the online iterative version so as to make better use of the target domain data in an unsupervised manner. The
iterative back-translation scheme we adopt has achieved great success in unsupervised NMT and text style transfer in the past two years~\cite{he2016dual,Lample2018PhraseBasedN,artetxe2017unsupervised,DBLP:conf/emnlp/JinJMMS19}. In this paper, we propose a novel adaptation of it to the specific setting of semi-supervised DAMT and achieve profound improvements over all previous state-of-the-art methods.

\section{Method}

In this section, we first introduce the architecture of our method, and then formulate the overall training strategy.

\subsection{Model Architecture}

\begin{figure}[!htpb]
    \centering
    \includegraphics[width= \columnwidth]{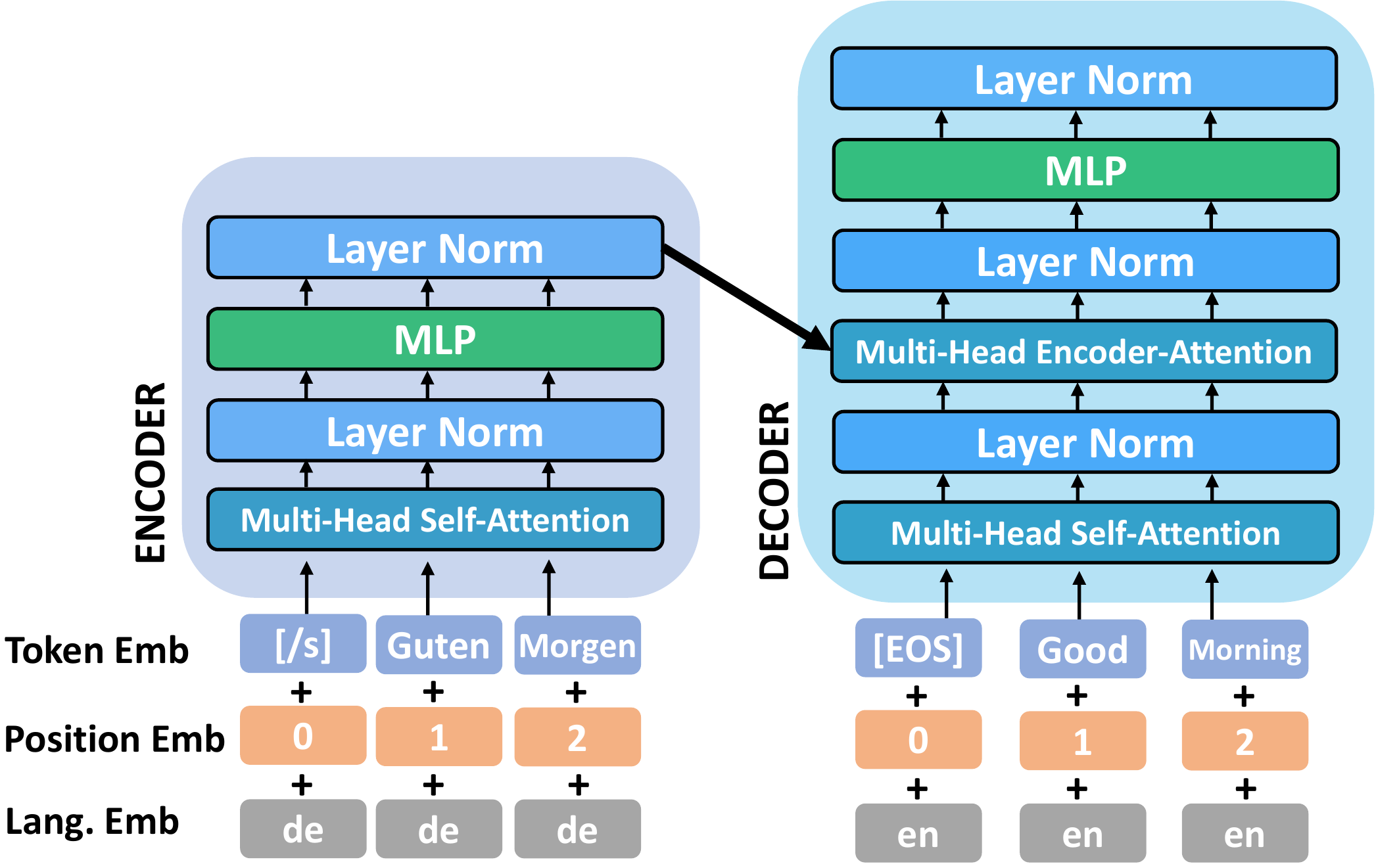}
    \caption{Transformer-based model architecture.}
    \label{fig:model}
\end{figure}

We adopt the Transformer~\cite{vaswani2017attention} with the encoder-decoder structure as the sequence-to-sequence translation model, as shown in Figure~\ref{fig:model}. Following the practice in ~\cite{Lample2019CrosslingualLM}, we add the language embeddings to the standard token and position embeddings via the element-wise summation operation. This language embedding can inform both encoder and decoder which language it is processing. For instance, when translating from German to English, we set the language embedding of the encoder to German (through a look-up table) while setting that of the decoder to English. For the reversed direction of translation (i.e., from English to German), we just need to reverse the language embedding settings for the encoder and decoder without changing the model architecture. In this way, the same model can be used to translate any language pair. 

Three key properties of our model are introduced in the following paragraphs:

\paragraph{Shared Sub-Word Vocabulary}
In our experiments, we process all languages with the same shared vocabulary created through Byte Pair Encoding (BPE)~\cite{sennrich-etal-2016-neural}. This not only enables us to translate between any language pair with the same model, but improves the alignment of embedding spaces across languages that share either the same alphabet or anchor tokens such as digits. The BPE splits are learned on the concatenation of sentences from the monolingual corpora.

\paragraph{Shared Latent Representations}
All encoder parameters (including the embedding matrices, since we perform joint tokenization) are shared across the source and target languages so that the encoder can map the input of any source language into a shared latent representation space, which is then translated to the target language by the decoder. Furthermore, we share the decoder parameters across the two languages to reduce parameter size. We also share the encoder and decoder between the translation and language modeling tasks (will be introduced in the Section \ref{sec:training-objectives}), which ensures that the benefits of language modeling, implemented via the denoising auto-encoder objective, nicely transfer to translation and helps the NMT model translate more fluently.

\paragraph{Pre-Training}
Both the encoder and decoder are initialized by pre-trained parameters from ~\citet{Lample2019CrosslingualLM}, which are obtained by pre-training a transformer based language model with both the conditional language modeling (CLM) and masked language modeling (MLM) objectives on large-scale monolingual corpora of both languages in the language pair (the corpora are extracted from the Wikipedia dump). We refer readers to the original paper for more technical details on the pre-training process. Such initialization not only accelerates the model convergence but also improves the adaptation performance, which will be discussed in Section~\ref{sec:init-useful}.

\subsection{Training Objectives}
\label{sec:training-objectives}

In the semi-supervised domain adaptation setting, we assume access to \textit{parallel} translation pairs as the training corpus $(X_{\mathrm{src}}, Y_{\mathrm{src}})$ in the source domain, and \textit{non-parallel} data $X_{\mathrm{tgt}}$ and $Y_{\mathrm{tgt}}$ in the target domain. As illustrated by Figure \ref{fig:method-schema}, we train our model with the following three objectives:

\paragraph{Language Modeling (LM)}
The language modeling objective is implemented via denoising auto-encoding (DAE),
by minimizing
\begin{align}
\label{eq:lm-los}
    \mathcal{L}_{\mathrm{lm}}=&\mathbb{E}_{\bm{x}\in X_{\mathrm{tgt}}}[-\log P_{\mathrm{s}\rightarrow \mathrm{s}}(\bm{x}|C(\bm{x});\bm{\theta})]+\nonumber\\
    & \mathbb{E}_{\bm{y}\in Y_{\mathrm{tgt}}}[-\log P_{\mathrm{t}\rightarrow \mathrm{t}}(\bm{y}|C(\bm{y});\bm{\theta})]
    \text{,}
\end{align}
where $C$ is a word corruption function with some words randomly dropped, blanked, and swapped; $P_{\mathrm{s}\rightarrow \mathrm{s}}$ and $P_{\mathrm{t}\rightarrow \mathrm{t}}$ are the composition of encoder and decoder both operating on the source language $s$ and target language $t$, respectively; $\theta$ denotes the model parameters.

\paragraph{Iterative Back-Translation (IBT)}
We have two NMT models, $\mathrm{Model}_{\mathrm{s2t}}(\cdot)$ which translates from the source language $s$ to the target language $t$, and $\mathrm{Model}_{\mathrm{t2s}}(\cdot)$ vice versa (they are implemented by one model architecture). In each iteration, we translate on the fly from each source language sentence in the target domain $\bm{x}\in X_{\mathrm{tgt}}$ to the target language sentence $\mathrm{Model}_{\mathrm{s2t}}(\bm{x})$.
Similarly, we translate from every target sentence $\bm{y}\in Y_{\mathrm{tgt}}$ to its counterpart in the source language $\mathrm{Model}_{\mathrm{t2s}}(\bm{y})$.
Then the pairs of $(\bm{x}, \mathrm{Model}_{\mathrm{s2t}}(\bm{x}))$ and $(\mathrm{Model}_{\mathrm{t2s}}(\bm{y}), \bm{y})$ can be used as synthetic parallel data to train the NMT model in two directions by minimizing the following loss:
\begin{align}
\label{eq:back-los}
    \mathcal{L}_{\mathrm{back}}=&\mathbb{E}_{\bm{x}\in X_{\mathrm{tgt}}}[-\log P_{\mathrm{t}\rightarrow \mathrm{s}}(\bm{x}|\mathrm{Model}_{\mathrm{s2t}}(\bm{x});\bm{\theta})]+\nonumber\\
    & \mathbb{E}_{\bm{y}\in Y_{\mathrm{tgt}}}[-\log P_{\mathrm{s}\rightarrow \mathrm{t}}(\bm{y}|\mathrm{Model}_{\mathrm{t2s}}(\bm{y});\bm{\theta})]
    \text{.}
\end{align}

To be noted, when minimizing this objective function, we do not back-propagate through the models that are used to generate translations.

\paragraph{Supervised Machine Translation}
When given parallel data, denoted as $(X_{\mathrm{para}}, Y_{\mathrm{para}})$, we can also minimize the supervised translation loss:
\begin{equation}
\label{eq:sup-los}
    \mathcal{L}_{\mathrm{sup}}=\mathbb{E}_{\bm{x}\in X_{\mathrm{para}},\bm{y}\in Y_{\mathrm{para}}}[-\log P_{\mathrm{s}\rightarrow \mathrm{t}}(\bm{y}|\bm{x};\bm{\theta})].
\end{equation}

The parallel data can be the source domain supervised data $(X_{\mathrm{src}}, Y_{\mathrm{src}})$ or the back-translated synthetic pairs by an NMT model trained on the source domain data.

\subsection{Training Strategy}

As shown in Algorithm \ref{alg:training-strategy}, in each iteration, we randomly draw a batch of data to minimize the aforementioned three loss equations \ref{eq:lm-los}, \ref{eq:back-los} and \ref{eq:sup-los}. The training will continue until the validation set BLEU score does not increase for a certain number of iterations.

\begin{algorithm}[t]
\small
\caption{Training Strategy}
\label{alg:training-strategy}
\begin{algorithmic}[1]
\REQUIRE{Non-parallel data $X_{\mathrm{tgt}}$ and $Y_{\mathrm{tgt}}$ in the target domain, parallel data $(X_{\mathrm{para}}, Y_{\mathrm{para}})$, and model parameters $\bm{\theta}$}
\WHILE{$\bm{\theta}$ has not converged} 
\STATE{Sample $\bm{x}$ from $X_{\mathrm{tgt}}$ and $\bm{y}$ from $Y_{\mathrm{tgt}}$}
\STATE{Create pairs $(C(\bm{x}), \bm{x})$ and $(C(\bm{y}), \bm{y})$ via word corruption}
\STATE{Update $\bm{\theta}$ by minimizing Eq.~\eqref{eq:lm-los}}
\STATE{Sample $\bm{x}$ from $X_{\mathrm{tgt}}$ and $\bm{y}$ from $Y_{\mathrm{tgt}}$}
\STATE{Create $(\bm{x}, \mathrm{Model}_{\mathrm{s2t}}(\bm{x}))$ and $(\mathrm{Model}_{\mathrm{t2s}}(\bm{y}), \bm{y})$ via back-translation}
\STATE{Update $\bm{\theta}$ by minimizing Eq.~\eqref{eq:back-los}}
\STATE{Sample $(\bm{x}, \bm{y})$ from $(X_{\mathrm{para}},Y_{\mathrm{para}})$}
\STATE{Update $\bm{\theta}$ by minimizing Eq.~\eqref{eq:sup-los}}
\ENDWHILE
\end{algorithmic}
\end{algorithm}

\section{Experiments}

\subsection{Datasets}

We validate our model under two different adaptation settings. For the first setting, we test the domain adaptation ability of our method for adapting from one specific domain to another specific one on every pair of the following specific domains: law (LAW),
medical (MED),
and Information Technology (IT).
The other setting is to adapt models trained on the general-domain WMT datasets to specific domains: TED~\cite{duh18multitarget}, LAW, and MED datasets~\cite{TIEDEMANN12.463}. Two language pairs are tested, German-English (DE-EN) and Romanian-English (RO-EN). The general-domain WMT datasets for DE-EN and RO-EN come from the WMT-14\footnote{https://www.statmt.org/wmt14/translation-task.html} and WMT-16\footnote{https://www.statmt.org/wmt16/translation-task.html} tasks, respectively. Data statistics for the train sets are in Table~\ref{tab:data-stats}. The size of validation and test sets for WMT-14 are both 3K, and all the other domains are 2K.

For fair comparison, we follow previous works~\cite{hu-etal-2019-domain,Dou2019UnsupervisedDA} to build the unaligned corpus for each domain. Specifically, we randomly shuffle the original parallel corpus and split it into two equal halves. We then use the first half of the source-side sentences and the second half of the target-side ones, which form the non-parallel corpus for the target domain.
In this way, we assure that there are no parallel sentences in the target domain. 

\begin{table}[!htbp]
\resizebox{0.48\textwidth}{!}{\begin{tabular}{llrrr}
\toprule
\textbf{Lang.}                  & \textbf{Corpus} & \textbf{Words}     & \textbf{Sentences} & \textbf{W/S}   \\ \midrule
\multirow{5}{*}{De-En} & MED   & 14,533,613  & 1,104,752 & 13.2 \\
                       & LAW    & 18,461,140  & 715,372   & 25.8 \\
                       & IT     & 3,212,130   & 337,817    & 9.5  \\
                       & TED    & 3,110,970   & 151,627    & 20.5 \\
                       & WMT-14  & 126,735,962 & 4,468,840   & 28.4 \\ \midrule
\multirow{4}{*}{Ro-En} & MED   & 13,142,512  & 990,220   & 13.3 \\
                       & LAW    & 10,631,517  & 450,715   & 23.6 \\
                       & TED    & 3,328,621   & 161,291   & 20.6 \\
                       & WMT-16  & 10,796,138  & 399,375   & 27.0 \\
\bottomrule
\end{tabular}}
\caption{Statistics of the corpora used for training (target side).}
\label{tab:data-stats}
\end{table}

\subsection{Baselines}

We compare our models with the following baselines described below.

\paragraph{\textsc{Unadapted}}
We train the NMT model on the supervised source domain data and directly test its performance on the target domain.

\paragraph{\textsc{Copy}~\cite{Currey2017CopiedMD}}
The target-side sentences in the target domain are copied to the source-side, and then they are combined with the parallel source domain data as the train data to train an NMT model.

\paragraph{\textsc{Back}~\cite{sennrich2015improving}}
This is the one time \textit{back-translation} baseline. A target-to-source NMT model is first trained on the parallel source domain data and then used to generate pseudo parallel data in the target domain for model training by translating the target domain target-side sentences to the source side.  

\paragraph{\textsc{DALI}~\cite{hu-etal-2019-domain}}
Lexicon induction is first performed to extract a lexicon in the target domain, and then a pseudo-parallel target domain corpus is constructed by performing word-to-word back-translation of monolingual sentences of the target language, which is used for fine-tuning a pre-trained source domain NMT model.

\paragraph{\textsc{DAFE}~\cite{Dou2019UnsupervisedDA}}
It performs multi-task learning on a translation model on source domain parallel data and a language model on target domain target-side monolingual data, while inserting domain and task embedding learners into the transformer-based model.


\subsection{Settings}

Both encoder and decoder in the transformer model have $6$ layers, $8$ heads, and a dimension of $1024$. For the word corruption function, word dropping and blanking adopt a uniform distribution with a probability of $0.1$, and word shuffling is implemented with a window of $3$ tokens. The Adam optimizer uses a learning rate of $0.0001$.

Our implemented methods involve three variants:

\paragraph{\textsc{\MethodName}}
It serves as the base of our method. For this variant, we do not use any supervised data. And we train our model by optimizing Equation \ref{eq:lm-los} and \ref{eq:back-los} only with the target domain non-parallel data.

\paragraph{\textsc{\MethodName+Src}}
Based on the variant of \textsc{\MethodName}, besides optimizing Equation \ref{eq:lm-los}\&\ref{eq:back-los}, we additionally optimize Equation \ref{eq:sup-los} using the supervised data from the source domain.

\paragraph{\textsc{\MethodName+Back}}
Similar to the variant of \textsc{\MethodName+Src}, instead of using the source domain data for solving Equation \ref{eq:sup-los}, we use the pseudo parallel data provided by the aforementioned baseline \textsc{Back}.

\begin{table*}[t]
\setlength{\tabcolsep}{7pt} 
\renewcommand{\arraystretch}{1.1} 
\resizebox{\textwidth}{!}{\begin{tabular}{ll|cc|cc|cc|ccc|ccc}
& \multirow{3}{*}{\textbf{Methods}} & \multicolumn{9}{c|}{\textbf{DE to EN}}                                                                            & \multicolumn{3}{c}{\textbf{RO to EN}} \\ \cline{3-14}
                         & & \multicolumn{2}{c|}{\textbf{LAW}} & \multicolumn{2}{c|}{\textbf{MED}} & \multicolumn{2}{c|}{\textbf{IT}} & \multicolumn{3}{c|}{\textbf{WMT-14}} & \multicolumn{3}{c}{\textbf{WMT-16}}   \\ \cline{3-14}
                         & & \textbf{MED}        & \textbf{IT}         & \textbf{LAW}        & \textbf{IT}         & \textbf{LAW}        & \textbf{MED}       & \textbf{TED}     & \textbf{LAW}     & \textbf{MED}    & \textbf{TED}      & \textbf{LAW}     & \textbf{MED}     \\ \hline\hline 
(1) & \textsc{Unadapted}            & 18.76      & 6.62       & 7.92       & 5.94       & 6.19       & 10.90     & 23.36   & 23.77   & 24.42  & 23.59    & 33.26   & 18.39   \\ \hline
(2) & \textsc{Copy}                 & 23.57      & 10.58      & 11.44      & 12.83      & 9.39       & 18.19     & 24.32   & 25.25   & 27.67  & 29.29    & 38.23   & 27.37   \\
(3) & \textsc{Back}                 & 33.94      & 22.21      & 23.74      & 23.56      & 22.43      & 31.00     & 31.02   & 31.27   & 35.69  & 36.98    & 49.28   & 43.70   \\
(4) & \textsc{DALI}                 & 11.32      & 8.75       & 26.98      & 19.49      & 11.65      & 10.99     & --     & --     & --    & --      & --     & --     \\
(5) & \textsc{DAFE}                 & 26.96      & 15.41      & 14.28      & 13.03      & 11.67      & 21.30     & \textbf{34.89}   & 31.46   & 38.79  &  37.05${}^\dagger$        &  49.63${}^\dagger$       &   46.77${}^\dagger$      \\ \hline
(6) & \MethodName                 & 38.67      & 31.69      & 27.89      & 31.69      & 27.89      & 38.67     & 30.88   & 27.89   & 38.67  & 34.48    & 49.45   & 61.55   \\
(7) & \textsc{\MethodName+Src}           & \textbf{41.22}      & 34.33      & 29.54      & 32.47      & 30.20      & 39.77     & 33.23   &   32.81      &  41.40      & 38.68    &    53.49        & 60.98         \\
(8) & \textsc{\MethodName+Back}          & 40.40      & \textbf{35.41}      & \textbf{30.27}      & \textbf{35.76}      & \textbf{30.49}      & \textbf{40.28}     & 34.15   & \textbf{33.35}   & \textbf{42.08}  & \textbf{38.90}    & \textbf{54.39}   & \textbf{66.08}   \\ \hline
(9) & MT (Sup.)          & 48.95      & 59.38      & 37.72      & 59.38      & 37.72      & 48.95     & 38.97   & 37.72   & 48.95  & 42.22    & 61.69   & 80.32  \\ 
\end{tabular}}
\caption{Translation accuracy (BLEU) under different settings. The second and third columns are source and target domains respectively. ``DE'', ``EN'', and ``RO'' denote German, English, and Romanian, respectively. DALI and DAFE results are the best results from the original papers, except that numbers marked by $\dagger$ are from our re-implementation. Settings (7) \textsc{\MethodName+Src} and (8) \textsc{\MethodName+Back} uses the out-of-domain data and back-translated data to minimize the supervised machine translation loss, respectively. (9) ``MT (Sup.)'' results are obtained by training an NMT model on the supervised target domain data.}
\label{tab:main-results}
\end{table*}

\section{Results}

\subsection{Main Results}

\paragraph{Adapting between Specific Domains}
Our main results are shown in Table~\ref{tab:main-results}, with the left six columns showing the adaptation setting where models are adapted between specific domains. In this table, the second row lists the source domains whereas the third row shows the target domains. From this table, we see that the unadapted baseline model, \textsc{Unadapted}, performs very poorly, verifying the previous statement that current NMT models cannot generalize well to test data from a new domain. In contrast, the copy method, \textsc{Copy}, and back-translation method, \textsc{Back}, can significantly improve the adaptation performance, with \textsc{Back} showing much superior performance. To be noted, \textsc{Back} even outperforms the other two baselines: \textsc{DALI} and \textsc{DAFE}, by a large margin in the majority of cases, although it was proposed earlier and is much simpler. 

Our method variant \textsc{\MethodName}, shown in row (6) of Table \ref{tab:main-results}, achieves higher performance than all baselines, with absolute gains of $+0.91$ to $+9.48$ BLEU scores over the strongest baseline, and $+19.91$ to $+27.77$ BLEU scores over the \textsc{Unadapted} baseline. Notably, {\MethodName} only needs the target domain non-parallel data but can still substantially outperform those baselines that rely on the parallel source domain data (e.g., \textsc{Back}, \textsc{DALI}, \textsc{DAFE}), indicating that previous methods have not exhausted the potential contained within the target domain data. 

\paragraph{Adapting from a General to a Specific Domain}
In the second adaptation setting, the right six columns of Table~\ref{tab:main-results} show the results by adapting a NMT model trained on the general domain corpus (WMT) to specific domains (TED, LAW, and MED) for two language pairs: from German to English and from Romanian to English. In this setting, both \textsc{Copy} and \textsc{Back} achieve better performance compared to the previous setting where the source domain is specific. 
Our method variant \textsc{\MethodName} surpasses \textsc{Unadapted} by a large margin but it does not outperform the baselines \textsc{Back} and \textsc{DAFE} in some cases. To complement this gap, we next augment \textsc{\MethodName} with supervised data either directly from the source domain or from the back-translated data using a NMT model trained on the source domain.

\paragraph{Combining \textsc{\MethodName} with Source Domain or Back-Translated Data}
\textsc{\MethodName} is trained purely on the non-parallel target domain data and can be augmented by adding supervised data, which leads to the two variants: \textsc{\MethodName+Src} and \textsc{\MethodName+Back}.
In row (7) of Table~\ref{tab:main-results}, for \textsc{\MethodName+Src}, we insert the supervised translation task using the source domain data, which can bring in around $+1$ to $+4$ BLEU improvements consistently compared with \textsc{\MethodName} (except for the MED target domain in the ro-en language pair). For \textsc{\MethodName+Back}, we replace the source domain data with the back-translated data provided by the baseline \textsc{Back},
and it achieves even better performance, as shown in row (8) of Table~\ref{tab:main-results}. The superior performance of both variants demonstrates the benefit from the supervised data. And by comparing \textsc{\MethodName+Src} and \textsc{\MethodName+Back}, we see that although the back-translated data is obtained by performing inference of a model trained on the source domain data, the back-translated data is still a better option for domain adaptation than the latter one.
Overall, our best setting can harvest up to $+19.31$ BLEU improvement over the best baseline model and $+47.69$ BLEU improvement over the \textsc{Unadapted} model. Notably, we have also tried combining the source domain parallel data with the back-translated data for supervised translation training but found that it performs worse than current settings. 

\subsection{Ablation Study}

To check the importance of the each component in our model, we conduct an ablation study on the domain adaptation performance of the best performing model, \textsc{\MethodName+Back}. We report the validation set BLEU scores by adapting from the LAW domain to two target domains, MED and IT, in Table~\ref{tab:ablation}. 
\begin{table}[!htpb]
\centering
\resizebox{\columnwidth}{!}{
\begin{tabular}{lcc}
\toprule
\multirow{2}{*}{\textbf{Model}} & \multicolumn{2}{c}{\textbf{Target Domain BLEU}} \\ 
& \textbf{MED}        & \textbf{IT}         \\ \midrule
\textsc{{\MethodName}+Back}                       & \textbf{42.13}      & \textbf{47.64}      \\
-- Pre-training         & 31.80 ($\downarrow$10.33)     & 27.71 ($\downarrow$19.93)      \\
-- $\mathcal{L}_{\mathrm{bt}}$                & 33.75 ($\downarrow$8.38)      & 25.37 ($\downarrow$22.27)      \\
-- $\mathcal{L}_{\mathrm{lm}}$           & 40.97 ($\downarrow$1.16)      & 40.82 ($\downarrow$6.82)      \\
-- Source-side $\mathcal{L}_{\mathrm{lm}}$       & 40.04 ($\downarrow$2.09)      & 42.66 ($\downarrow$4.98)      \\
-- $\mathcal{L}_{\mathrm{bt}}$ -- Source-side $\mathcal{L}_{\mathrm{lm}}$     & 37.29 ($\downarrow$4.84)      & 35.06 ($\downarrow$12.58)     \\ \bottomrule
\end{tabular}
}
\caption{Ablation study on the domain adaptation performance of \textsc{\MethodName+Back}. The source domain is LAW, and target domains are MED and IT. 
``-- Source-side $\mathcal{L}_{\mathrm{lm}}$'' means no language modeling on the source-side sentences.
}
\label{tab:ablation}
\end{table}

The most important components of our model are \textit{Pre-training} and $\mathcal{L}_{\mathrm{bt}}$. If the model is not initialized with pre-trained parameters (``-- Pre-training''), the model suffers from a substantial performance decrease by 10 to 20 BLEU scores. If we remove the iterative back translation objective ($\mathcal{L}_{\mathrm{bt}}$), the performance also drops significantly, $\downarrow$8.38 on MED and $\downarrow$22.27 on IT. We also find that $\mathcal{L}_{\mathrm{lm}}$ and source-side $\mathcal{L}_{\mathrm{lm}}$ are also important to the BLEU scores but not as crucial as the previous two components.
However, an interesting finding is that if back translation $\mathcal{L}_{\mathrm{bt}}$ and the source-side language modeling $\mathcal{L}_{\mathrm{lm}}$ are removed together (``-- $\mathcal{L}_{\mathrm{bt}}$ -- Source-side $\mathcal{L}_{\mathrm{lm}}$''), the domain adaptation performance is better than mere ``-- $\mathcal{L}_{\mathrm{bt}}$''. The reason is that if back translation is removed, the decoder just need to learn the target language, and language modeling on the source-side will impose a negative effect.

\section{Discussion}

\subsection{Is Pre-Training Always Helpful?} \label{sec:init-useful}

Through the experiments, we find that initializing the translation model with pre-trained parameters can not only benefit our method but also the baselines. In Table~\ref{tab:pre-trainin}, we compare two settings: with and without pre-training, for three baselines: \textsc{Unadapted}, \textsc{Copy}, \textsc{Back}, where we adapt from the LAW domain to MED and IT domains. For all three baselines, pre-training consistently brings in substantial improvements, although the pre-training process is performed via unsupervised language modeling training on Wikipedia text that is irrelevant to the target domain data we used here. This shows that proper initialization of models is crucial to the domain adaptation problem to circumvent the lack of supervised data in the target domain. 

\begin{table}[!htpb]
\setlength{\tabcolsep}{7pt} 
\renewcommand{\arraystretch}{1.1} 
\resizebox{0.49\textwidth}{!}{\begin{tabular}{lcccccc}
\toprule
 \multirow{2}{*}{\textbf{Target}} & \multicolumn{2}{c}{\textbf{\textsc{Unadapted}}} & \multicolumn{2}{c}{\textbf{\textsc{Copy}}} & \multicolumn{2}{c}{\textbf{\textsc{Back}}} \\ 
                                                & \textbf{w/}             & \textbf{w/o}          & \textbf{w/}          & \textbf{w/o}        & \textbf{w/}          & \textbf{w/o}        \\ \midrule
 \textbf{MED}                     &\textbf{17.44}          &   9.69           & \textbf{24.38}       & 17.42      & \textbf{36.61}       & 26.75      \\
                         \textbf{IT}                      & \textbf{5.35}           &   3.87           & \textbf{8.73}        & 5.07       & \textbf{28.32}       & 9.21  \\ \bottomrule    
\end{tabular}}
\caption{The comparison of three baselines: \textsc{unadapted}, \textsc{copy}, and \textsc{back}, between cases with and without pre-training when adapting them from the domain of LAW to MED and IT. Validation set BLEU scores are reported.}
\label{tab:pre-trainin}
\end{table}

\subsection{Do More Non-parallel Data Help?}

One advantage of our method is that it keeps gaining improvement if the non-parallel data get larger. To verify this statement, we collected additional non-parallel data, combined them with the original target domain data,\footnote{When combining the extra non-parallel data with the original target domain data, we always up-sample the original data via replication so that it can have the same size as the additional data.} and analyzed the performance difference before and after adding these extra data, as shown in Table~\ref{tab:add-mono-data}. Specifically, we studied the adaptation from the WMT data to TED for both DE-EN and RO-EN language pairs. We consider two sources of extra non-parallel data: source domain and target domain. The source domain data source can be the WMT data itself, whereas for the target domain, we collected an additional dataset of TED talks. After scraping all the TED talk web-pages\footnote{https://www.ted.com/} until the early January 2020, we extracted the transcripts in three languages, English, German and Romanian, and kept the unique TED talk identifier of the transcript.
Note that for any language pair, we made sure that the transcript of a TED talk only appeared once in either the source or the target side to avoid any parallel sentences.

From Table~\ref{tab:add-mono-data}, we see that in general adding extra non-parallel data to our method can always lead to better performance, and choosing those extra data that have the same data distribution as the target domain is optimal. However, if we could not get more non-parallel data from the target domain, those in the source domain that are more readily to be obtained may also potentially improve the adaptation performance. 
By adding extra data, our best setting achieved even higher BLEU scores that are very close to supervised translation performance, as shown in Table~\ref{tab:add-mono-data}. 
This set of experiments have shown the great potential of our method: we can always seek to collecting more non-parallel data to keep improving the adaptation performance.


\begin{table}[!htpb]
\setlength{\tabcolsep}{7pt} 
\renewcommand{\arraystretch}{1.15} 
\resizebox{\columnwidth}{!}{\begin{tabular}{llcc}
\toprule
\multirow{2}{*}{\textbf{Model}} & \multirow{2}{*}{\textbf{+ Data}} & \textbf{WMT14$\rightarrow$}       & \textbf{WMT16$\rightarrow$}       \\ 
&                                   & \textbf{TED (DE-EN)} & \textbf{TED (RO-EN)} \\ \midrule
\textsc{\MethodName}         & --                              & 30.88       & 34.48       \\ 
\textsc{\MethodName}         & WMT                               & 32.45       & 37.03       \\ 
\textsc{\MethodName}         & TED                               & \textbf{33.34}       & \textbf{39.01}       \\ \midrule
\textsc{{\MethodName}+Back}    & --                              & 34.15       & 38.90       \\ 
\textsc{{\MethodName}+Back}    & WMT                              & 34.74       & 38.79       \\
\textsc{{\MethodName}+Back}    & TED                               & \textbf{36.45}       & \textbf{40.92}       \\ \midrule
MT (Sup.)            & --                              & 38.97       & 42.22       \\ \bottomrule
\end{tabular}}
\caption{Test set BLEU scores after adding extra non-parallel data (``+Data'') from the source WMT domain (``WMT'') or from the target TED domain (``TED'').}
\label{tab:add-mono-data}
\end{table}

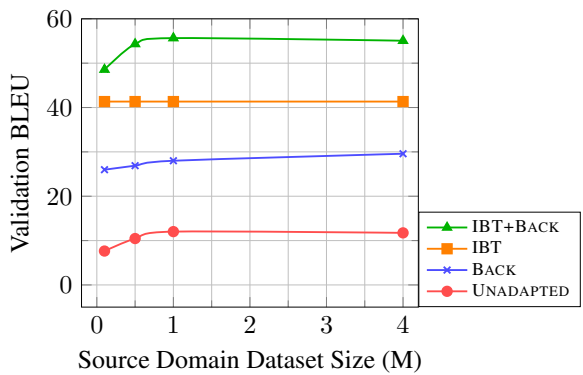
\begin{figure}[ht!]
  \centering
   \resizebox{\columnwidth}{!}{%
            \begin{tikzpicture}
        \pgfplotsset{
            scale only axis,
            xmin=-0.2, xmax=4.2,
            xtick={0,...,4.2},
            legend style={at={(1,0)},anchor=south west},
        }
        \begin{axis}[
          grid=both,
          legend style={cells={align=left},nodes={scale=0.7, transform shape},
          legend cell align={left}},
          ymin=-5, ymax=60,
          xlabel=Source Domain Dataset Size (M),
          ylabel=Validation BLEU,
          minor tick num=1,
          every axis plot/.append style={thick}
        ]
        \addplot[smooth,mark=triangle*,black!30!green]
          coordinates{
            (0.1, 48.55)
            (0.5, 54.32)
            (1, 55.65)
            (4, 55.06)
        }; \addlegendentry{\textsc{\MethodName+Back}}
        
        \addplot[smooth,mark=square*,orange]
          coordinates{
            (0.1, 41.34)
            (0.5, 41.34)
            (1, 41.34)
            (4, 41.34)
        };
        \addlegendentry{\MethodName{}}
        \addplot[smooth,mark=x,white!30!blue]
          coordinates{
            (0.1, 25.98)
            (0.5, 26.89)
            (1, 28.00)
            (4, 29.58)
        };
        \addlegendentry{\textsc{Back}}
        
        \addplot[smooth,mark=*,white!30!red]
          coordinates{
            (0.1, 7.64)
            (0.5, 10.46)
            (1, 12.02)
            (4, 11.74)
        };
        \addlegendentry{\textsc{Unadapted}}
        
        \end{axis}
        \end{tikzpicture}
        }
\caption{Effects of source domain dataset size on four adaptation methods: \textsc{Unadapted}, \textsc{Back}, \textsc{\MethodName}, and \textsc{{\MethodName}+Back}. We adapt from the general domain (WMT14) to the IT domain for the German-English language pair. All the target domain monolingual data, and sub-sampled 0.1, 0.5, 1, and 4 million source domain parallel pairs are used, respectively.}
\label{fig:semi}
\end{figure}

\subsection{How do Different Sizes of Source Domain Data Influence the Performance?}

In this section, we want to examine the effects of source domain dataset size on various adaptation methods.
To this end, we sub-sample the source domain parallel data at different sampling ratios, and report the validation set BLEU scores on four adaptation methods: \textsc{Unadapted}, \textsc{Back}, \textsc{\MethodName}, and \textsc{{\MethodName}+Back}, as shown in Figure~\ref{fig:semi}. Specifically, we adapt from the general domain (WMT14) to the IT domain for the German-English language pair. We use all the target domain non-parallel data, and sub-sample 0.1, 0.5, 1, and 4 million source domain parallel data. In Figure~\ref{fig:semi}, \textsc{\MethodName} does not use any source domain data so it stays unchanged, while all the other settings demonstrate improved performance with the increasing number of source domain data, and the performance gradually saturates when the number of source domain data exceed 1 million. Notably, \textsc{{\MethodName}+Back} consistently outperforms all others by a large margin, and its performance also increases at a higher rate, indicating that our method makes better use of the source domain supervised data. 


\section{Conclusion}
In this paper, we empirically identify that the iterative back-translation training scheme can yield large improvements to semi-supervised domain adaptation for NMT. On three low-resource domains, this basic approach demonstrates improvements of up to $+9.48$ BLEU scores over the strongest of four previous models, and up to $+27.77$ BLEU over the unadapted baseline. By further combining with popular data augmentation methods and utilizing supervised data from the source domain, our model shows a substantial improvement of up to $+19.31$ BLEU higher than the strongest baseline model, and up to $+47.69$ over the unadapted model. We put forward this method as a simple but strong baseline for semi-supervised domain adaptation for MT and future works in this direction should be compared with it.


\bibliographystyle{acl_natbib}
\bibliography{emnlp2020}




\end{document}